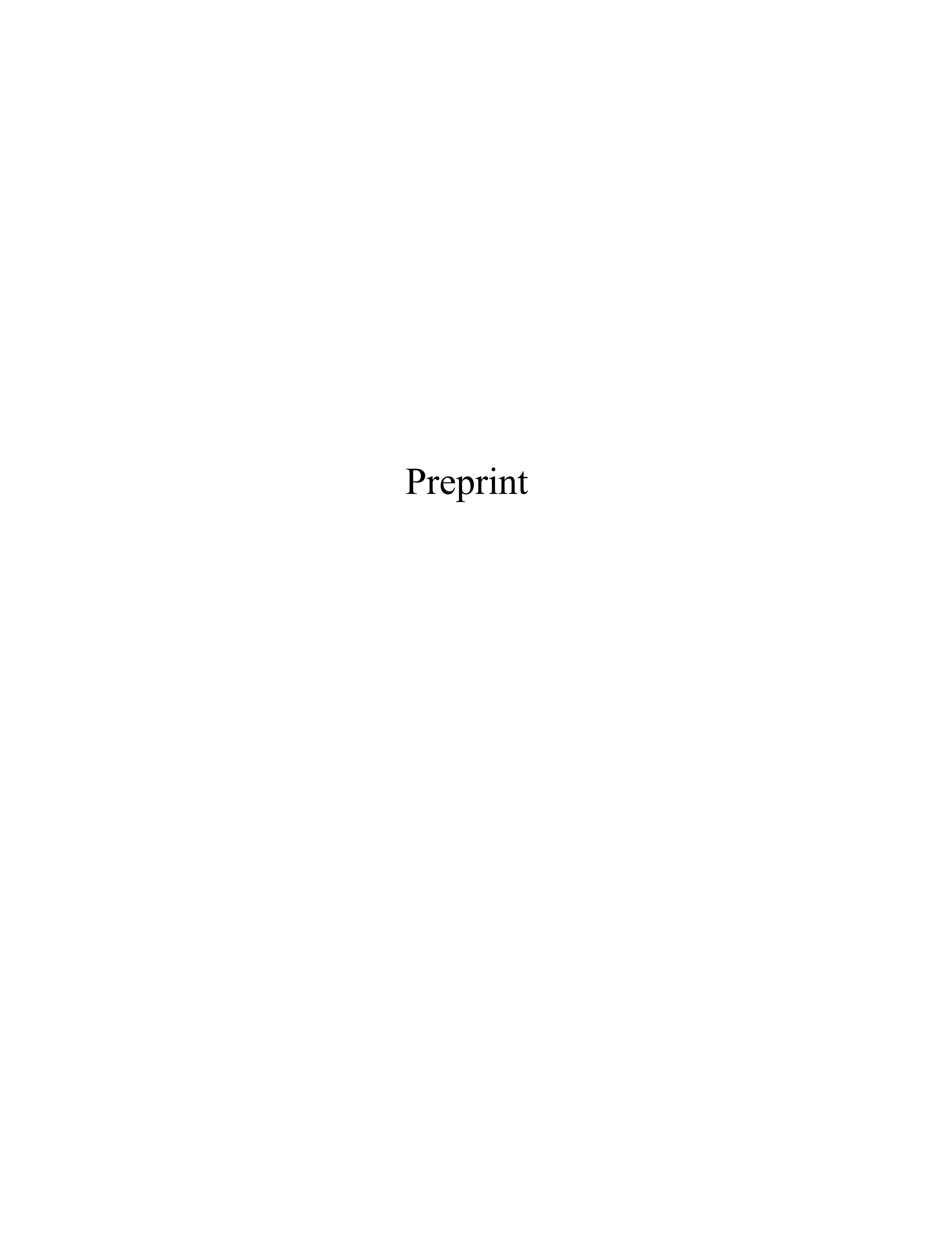

Preprint

# Credit Risk Identification in Supply Chains Using Generative Adversarial Networks

Zizhou Zhang
UIUC, Champaign, IL 61801, USA
zhangzizhou_2000@outlook.com

Xinshi Li
Montclair State University, Montclair, NJ 07043, USA
xinshili9@gmail.com

Yu Cheng
Columbia University, New York, NY 10027, USA
yucheng576@gmail.com

Zhenrui Chen*
Columbia University, New York, NY 10027, USA
*Corresponding author: zc2569@columbia.edu

Qianying Liu
Independent researcher, 731 Madison place, MA, 01772
liuqianying55@gmail.com

***Abstract*—Credit risk management within supply chains has emerged as a critical research area due to its significant implications for operational stability and financial sustainability. The intricate interdependencies among supply chain participants mean that credit risks can propagate across networks, with impacts varying by industry. This study explores the application of Generative Adversarial Networks (GANs) to enhance credit risk identification in supply chains. GANs enable the generation of synthetic credit risk scenarios, addressing challenges related to data scarcity and imbalanced datasets. By leveraging GAN-generated data, the model improves predictive accuracy while effectively capturing dynamic and temporal dependencies in supply chain data. The research focuses on three representative industries—manufacturing (steel), distribution (pharmaceuticals), and services (e-commerce)—to assess industry-specific credit risk contagion. Experimental results demonstrate that the GAN-based model outperforms traditional methods, including logistic regression, decision trees, and neural networks, achieving superior accuracy, recall, and F1 scores. The findings underscore the potential of GANs in proactive risk management, offering robust tools for mitigating financial disruptions in supply chains. Future research could expand the model by incorporating external market factors and supplier relationships to further enhance predictive capabilities.**

## I. INTRODUCTION

Supply chain risk management has consistently been a critical topic within the field of supply chain management, attracting considerable attention from both industry practitioners and academic researchers. As global market competition intensifies, the vulnerabilities within supply chain structures have become increasingly apparent. The intricate interconnections among supply chain entities—such as debt obligations, business transactions, and collateral guarantees—mean that the credit risk of one company can propagate to its directly linked partners, potentially affecting the entire supply chain network. Moreover, the impact of credit risk contagion varies significantly across different industries' supply chains [1]. Therefore, investigating the mechanisms of credit risk transmission within supply chains and the industry-specific differences in this contagion is of profound importance.

The ability to assess credit risk accurately is fundamental to the stability and sustainability of any supply chain. In particular, credit risk can arise from delays in payments, insolvency, or financial mismanagement, all of which threaten the liquidity and operational continuity of firms involved in the supply chain [2-3]. Traditional risk assessment methods, such as logistic regression and decision trees, are limited in their capacity to deal with dynamic and sequential supply chain data. These models tend to focus on static features and fail to capture complex, hidden patterns that evolve over time.

Recent advances in machine learning, especially Generative Adversarial Networks (GANs), have shown promise in solving these types of problems [4]. This paper introduces a GAN-based model to enhance credit risk identification in supply chains by simulating realistic credit risk events. The generative capability of GANs allows for the creation of more robust training datasets, enabling the model to learn better risk identification patterns. The aim of this study is to demonstrate that GANs can improve the accuracy and reliability of credit risk assessments in supply chains.

## II. LITERATURE REVIEW

The importance of credit risk management and its impact in supply chain finance has been well-documented in literature. Several approaches have been used to assess and mitigate credit risk, such as logistic regression, decision trees, and more recently, machine learning techniques like support vector machines (SVM) and random forests [5]. For example, Chen and Liu et al. (2024) developed a model with RPT network and DANE to generate condensed vector to generate condensed vector representations of the firms to predict the default distance [6]. While these traditional models offer certain advantages, they often fail to capture the temporal and sequential dependencies in transaction data, which are critical in understanding credit risk in dynamic supply chain environments [7-9].

In contrast, deep learning models, particularly Recurrent Neural Networks (RNNs) and Long Short-Term Memory (LSTM) networks, have gained significant traction due to their ability to handle sequential data [10-13]. However, these models still require large datasets for effective training, which is often a limitation in real-world applications [14-17]. Generative Adversarial Networks (GANs), first introduced by Goodfellow et al. (2014), have emerged as a powerful tool in data generation and augmentation [18-20]. Regarding credit risk, GANs have been used to generate synthetic financial data for training purposes. GANs excel in scenarios where real data is scarce or difficult to obtain, making them particularly suitable for supply chain risk prediction, where historical data on defaults or payment delays may be limited [21-24].

Recent research has also integrated sophisticated statistical techniques with advanced machine learning algorithms, yielding valuable new perspectives. For example, Ke and Yin (2024) demonstrated that a multivariate multilevel CAViaR model, optimized using gradient descent and genetic algorithms, provides a robust framework for analyzing credit risk and tail risk spillover within various sectors of the U.S. financial systems [25]. Their findings suggest that effective risk management, especially in credit risk-focused domains, does not necessarily require complex, computationally intensive models. Instead, by aligning the model with the specific application context and optimizing it appropriately, simpler models can lead to more efficient risk assessments. This approach highlights the potential of leveraging simpler, well-optimized models to achieve superior results without increasing model complexity.

While the use of GANs in financial applications is still in its nascent stages, several studies have highlighted their potential in fraud detection and credit scoring [26-28]. However, the specific application of GANs to credit risk identification within supply chains remains underexplored. This paper addresses this gap by proposing a GAN-based model tailored to supply chain credit risk detection [29-31].

## III. METHODOLOGY

This section outlines the methodology employed to develop and evaluate the GAN-based credit risk identification model for supply chains.

### A. *Data Collection*

Supply chains can generally be divided into two categories: product-based supply chains and service-based supply chains. The product-based supply chain is most widely used in industries such as manufacturing and distribution. Manufacturing is inherently linked to supply chain management; it is no exaggeration to state that the success or failure of supply chain management determines the fate of the manufacturing sector[32-35]. The steel industry, as the backbone of industrial development in developing countries, stands out as one of the most crucial manufacturing sectors. Its supply chain management evolution reflects the broader trends in supply chain management development within less developed nations. Furthermore, traditional distribution industries consist of two stages: wholesale and retail. As the distribution industry expands, it has gradually shifted away from manufacturing-centric supply chain management models and is now capable of taking the lead in supply chain operations. The pharmaceutical industry's supply chain model integrates elements of traditional distribution management while adopting characteristics of modern, distribution-led supply chain models. In the service sector, e-commerce has become the most complex and widely applied supply chain model[36]. Therefore, this study focuses on representative sectors across manufacturing, distribution, and service industries—namely, the steel industry, pharmaceutical distribution, and internet retailing—to examine the differential contagion of supply chain credit risk across these industries. Data sources include Wind, Bloomberg, and Reuters etc.

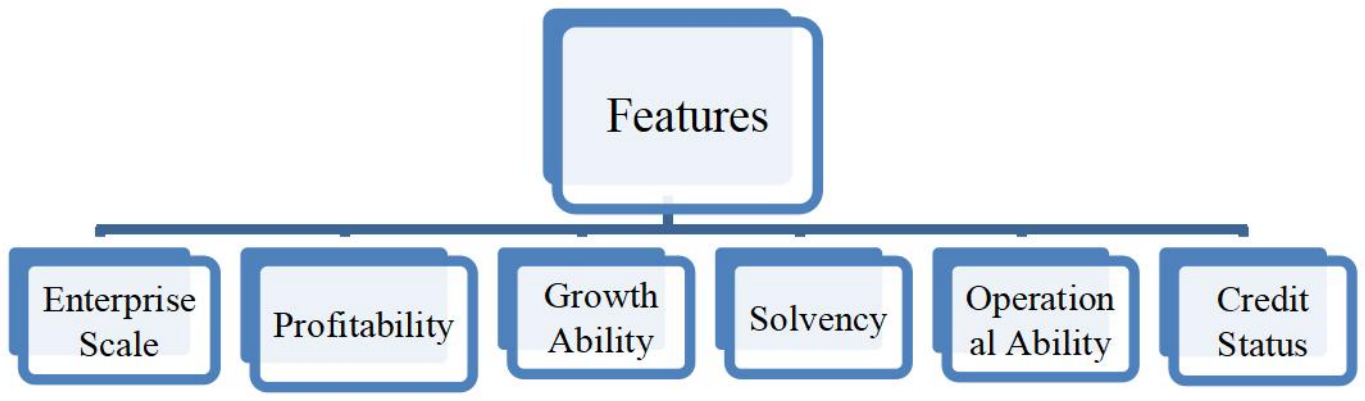

Figure 1. Variables set of the model

The evaluation indicators should more comprehensively consider the relationship between small and medium-sized enterprises (SMEs) and core enterprises, as well as the entire supply chain's dynamics, to better and more accurately assess the creditworthiness of financing SMEs[37-39]. Therefore, this study, from the perspective of supply chain finance, examines the financial and non-financial aspects of financing SMEs and core enterprises, along with the overall operational status of the supply chain. Building on the existing literature on supply chain finance indicator systems, this paper integrates both qualitative and quantitative indicators to redesign a credit risk evaluation system that reflects the risk level across the entire supply chain as Figure 1.

TABLE I. VRIABLE DEFINITIONS

| Category | Indicator Name |
|---|---|
| | Total Profit |
| | Operating Margin |
| **Profitability** | Capital Cost Profit Margin |
| | Return on Assets (ROA) |
| | Net Profit Growth Rate |
| | Total Assets |
| | Development Capability |
| **Assets and Growth** | Operating Revenue Growth Rate |
| | Total Asset Growth Rate |
| | Net Profit Growth Rate |
| **Liquidity Indicators** | Current Ratio |
| | Quick Ratio |
| | Inventory Turnover Rate |
| **Operational Efficiency** | Accounts Receivable Turnover Rate |
| | Total Asset Turnover Rate |
| **Contract Status** | Contract Status |

Specifically, as Table I, the key variables include total profit, total assets, operating margin, capital cost profit margin, return on assets, net profit growth rate, development capability, operating revenue growth rate, total asset growth rate, net profit growth rate, liquidity indicators such as current ratio and quick ratio, as well as operational efficiency metrics including inventory turnover rate, accounts receivable turnover rate, total asset turnover rate, and contract status.

## B. GANs Model Architecture

The GAN model consists of two primary components: the generator G and the discriminator D[40]. The generator creates synthetic credit risk scenarios based on the input features, while the discriminator evaluates the authenticity of the generated scenarios by distinguishing them from real data showed as Figure 2. The two networks are trained simultaneously, with the generator improving its ability to create realistic data and the discriminator becoming more adept at identifying fake data.

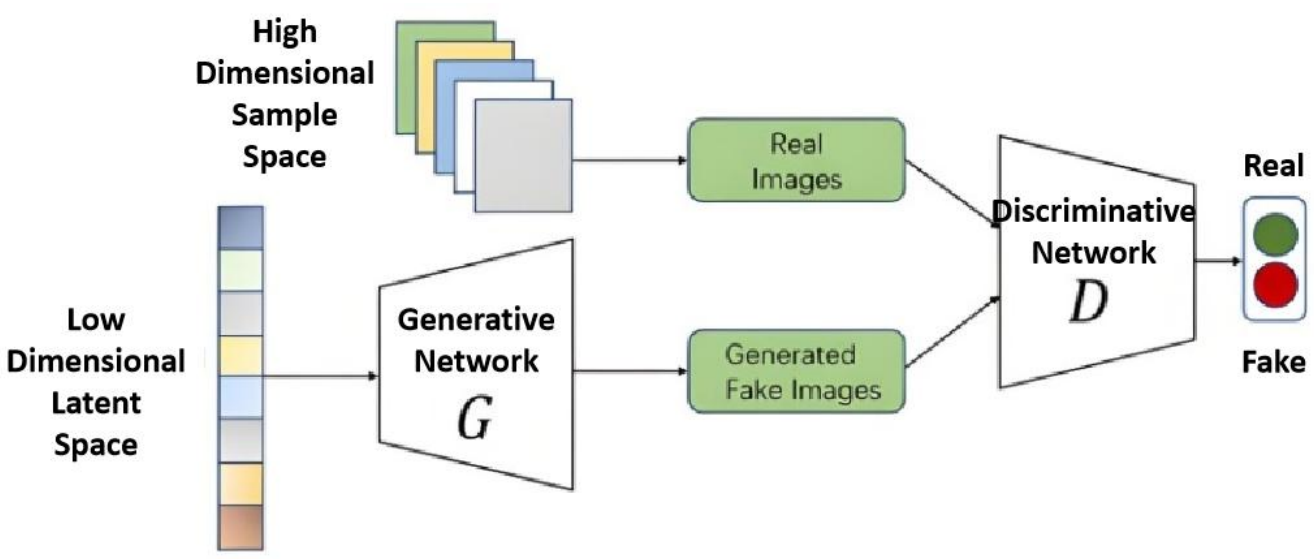

Figure 2. Process of GANs model

For this study, the generator is a multi-layer perceptron (MLP) with fully connected layers, designed to learn the complex relationships between the input features (e.g., transaction history, financial indicators) and the credit risk label (default or no default). The discriminator is also an MLP, designed to classify the generated data as either real or fake, based on its similarity to the actual dataset. Their objectives can be represented as a minimax optimization problem, typically expressed as:

$$\min_{G}\max_{D} V(D,G) = E_{x\sim P_{data}(x)}\left[\log\left(D(x)\right)\right] + E_{z\sim P_z(z)}[\log\left(1 - D(G(z))\right)] \quad (1)$$

where, $P_{data}(x)$ denotes the distribution of real data, P_z (z) is the distribution of the noise variable z, usually sampled from a uniform or normal distribution. G(z) represents the output of the generator, which produces synthetic samples, D(x) is the discriminator's output, indicating the probability that sample x is from the real data distribution. This formulation describes the generator's goal to minimize the discriminator's ability to distinguish between real and generated samples, while the discriminator aims to maximize its accuracy in this classification task.

To address the common issues of mode collapse, instability, and low training efficiency in GANs, this paper adopts the Wasserstein GAN (WGAN) framework. By utilizing the Wasserstein distance to compute the loss, WGAN mitigates the potential problem of gradient vanishing that often occurs in traditional GANs. Furthermore, WGAN incorporates weight clipping to enforce the Lipschitz constraint on the critic, which contributes to improved training stability. Additionally, batch normalization and label smoothing techniques are employed. Batch normalization is applied to each layer in both the generator and the discriminator, helping to stabilize and accelerate the training process while preventing gradient explosion or vanishing. Label smoothing, on the other hand, slightly smooths the real labels to prevent the discriminator from becoming overly confident, thereby enhancing training stability.

In terms of potential overfitting, Lai et al. (2024) designed GM-DF, an innovative regularization method that reduces conflicts during joint training on different datasets, thereby enhancing the model's adaptability and performance across multiple scenarios [41].

The GAN model is trained using the Adam optimizer, with a learning rate of 0.0002 and a batch size of 64. The training process is stopped once the generator produces sufficiently realistic data, as determined by the discriminator's classification accuracy.

## C. Model Evaluation

The evaluation metrics include accuracy, precision, recall, and F1-score. These metrics are computed on a test set that contains both real and synthetic data, ensuring that the model is evaluated on its ability to generalize to new, unseen credit risk events.

# IV. Results and Discussion

## A. Model Performance

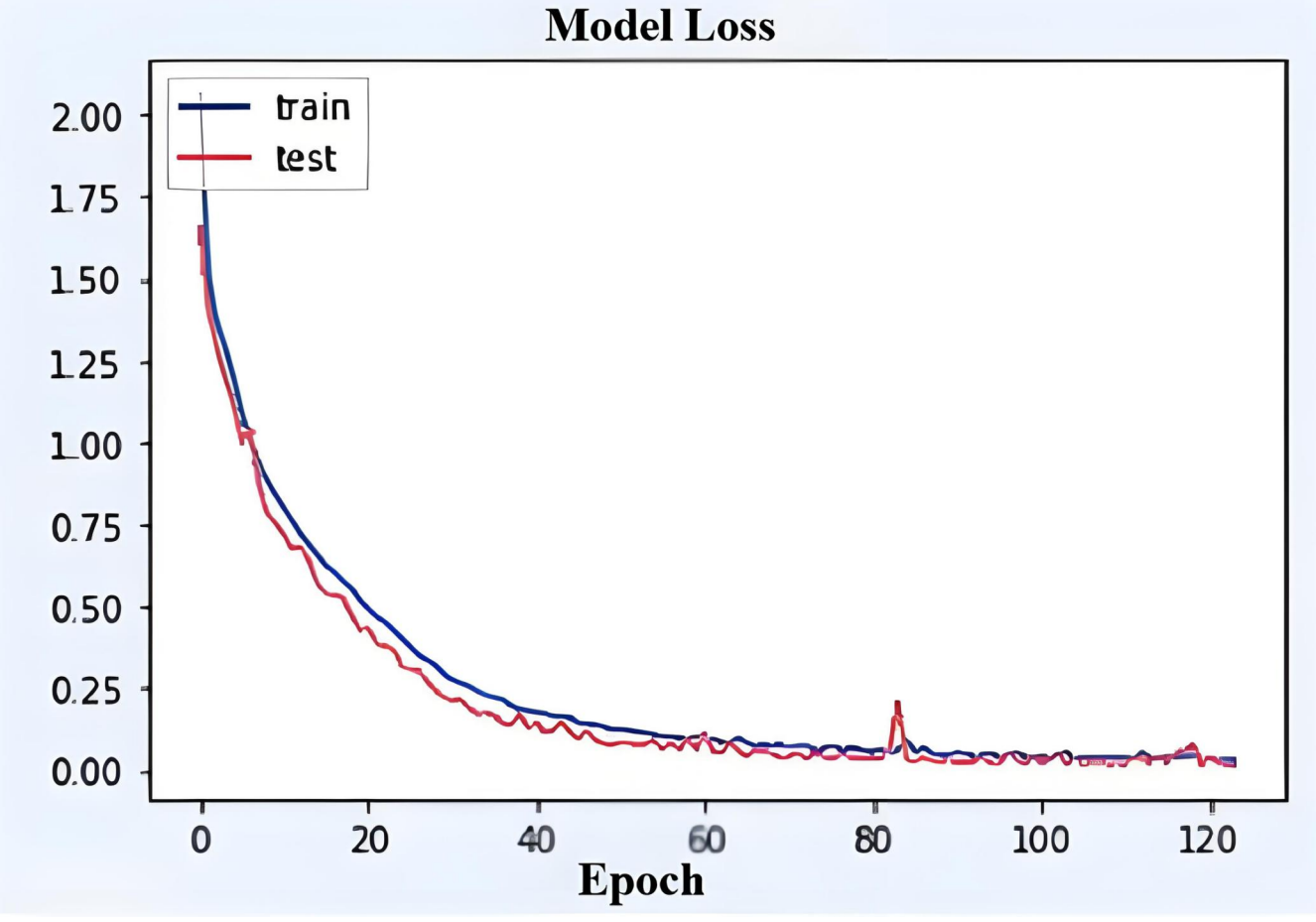

Figure 3. Loss function results

The Figure 3 illustrates the changes in the loss functions of a GAN model during the training process, comparing the loss for both the training set (blue curve) and the test set (red curve) over 120 epochs. Initially, both losses are high, reflecting the model's untrained state. As training progresses, the losses decrease rapidly, indicating that the generator and discriminator are improving their respective performances. Eventually, the losses stabilize at lower values, signifying that the model has reached a state of equilibrium. The minor fluctuations observed in later epochs are common in GAN training due to the dynamic interaction between the generator and discriminator.

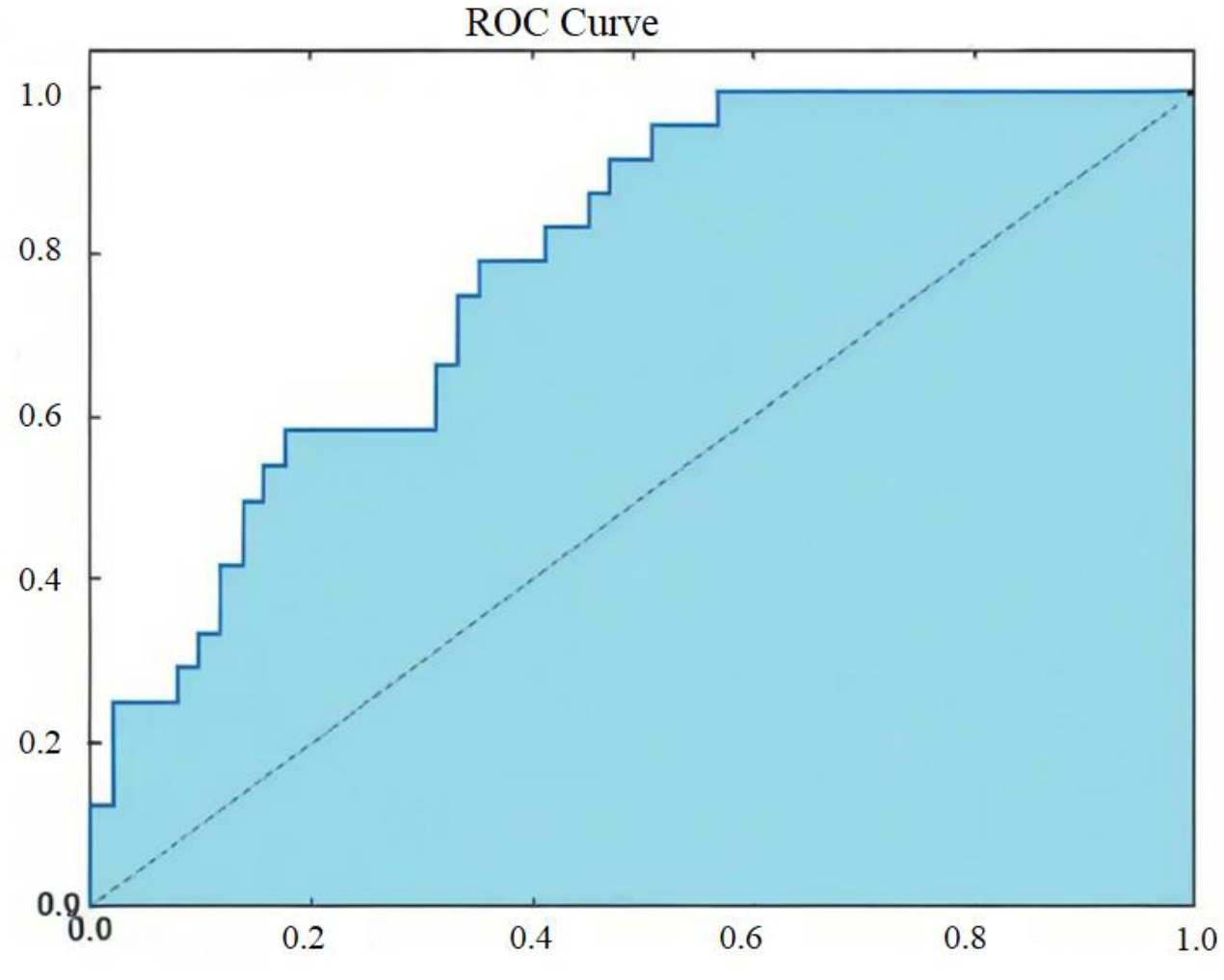

Figure 4. ROC Curve of the model

The Figure 4 displays the Receiver Operating Characteristic (ROC) curve for the GAN model, which evaluates its classification performance. The x-axis represents the false positive rate (FPR), while the y-axis indicates the true positive rate (TPR). The curve demonstrates the trade-off between sensitivity and specificity at various threshold settings. The area under the curve (AUC) serves as a quantitative measure of the model's ability to distinguish between classes. A curve closer to the top-left corner signifies higher accuracy, whereas the diagonal dashed line represents the baseline performance of a random classifier. The shaded area highlights the AUC, indicating the model's effectiveness in achieving a favorable balance between true positives and false positives.

### B. *Comparison of Models*

To further assess the performance of the GAN-based model, we compare its credit risk prediction capabilities with several baseline models, including logistic regression, decision trees, and support vector machines (SVM).

TABLE II. PERFORMANCE OF DIFFERENT MODELS

| Model | Accuracy | Recall | Precision | F1 Score |
|---|---|---|---|---|
| SVM | 0.83 | 0.88 | 0.84 | 0.89 |
| BP network | 0.88 | 0.93 | 0.88 | 0.94 |
| RNN | 0.9 | 0.96 | 0.9 | 0.95 |
| LSTM | 0.92 | 0.97 | 0.93 | 0.96 |
| GANs | 0.96 | 1 | 0.97 | 0.97 |

As Table II above, using the same dataset, we applied these models, and the results are summarized in the table above. Notably, GANs still demonstrated the most promising performance, while RNN LSTM also yielded favorable results. This indicates that neural networks, particularly those capable of capturing complex temporal dependencies like LSTM, are better suited for handling nonlinear problems in this context. The superior performance of GANs, in particular, suggests that generative models may offer significant advantages in modeling the underlying data distribution and producing high-quality predictions.

One of the key advantages of the GAN model is its ability to generate synthetic credit risk data, which augments the training dataset and improves the robustness of the model. The inclusion of synthetic data significantly enhances the performance of the model, especially in cases where historical credit risk events are sparse. When the GAN-generated data is removed, the performance of the model drops by approximately 5%, highlighting the importance of synthetic data in improving predictive accuracy.

### C. *Implications for Supply Chain Credit Risk*

The results demonstrate that the GAN-based approach can effectively identify credit risks in supply chains, even in the presence of limited or imbalanced data. This capability is particularly valuable in real-world applications where transaction data may be incomplete or where firms are hesitant to share sensitive financial information. The use of GANs can also lead to more proactive risk management strategies by providing early warnings of potential defaults, which can help firms mitigate losses and take preventive actions.

## V. CONCLUSIONS

This study advances the understanding of credit risk identification in supply chains by leveraging Generative Adversarial Networks (GANs). The application of GANs has demonstrated significant potential in enhancing predictive accuracy, addressing the challenges of data scarcity, and capturing complex dynamic dependencies within supply chain data. By generating synthetic credit risk scenarios, the model offers a robust framework that improves upon traditional methods such as logistic regression, decision trees, and neural networks.

The research highlights the effectiveness of GANs across different industries, including manufacturing, distribution, and services, showcasing the model's adaptability to varying risk propagation patterns. The empirical results indicate that the GAN-based approach outperforms existing models in key performance metrics, providing a more nuanced and proactive method for credit risk management.

Moreover, the study underscores the practical implications of using GANs in real-world supply chains, where data limitations often hinder traditional risk assessment models. The ability to generate high-quality synthetic data enables better training and ultimately more reliable risk predictions, offering firms a strategic tool to mitigate potential financial disruptions.

Looking forward, future research could integrate additional external factors such as macroeconomic indicators, market trends, and supplier relationship dynamics to further refine the predictive capabilities of the model. Expanding the dataset to include diverse global supply chain scenarios could also enhance the generalizability of the findings. The convergence of GANs with other advanced machine learning techniques promises a new frontier in supply chain finance, driving more resilient and adaptive risk management strategies.